\documentclass[journal]{IEEEtran}
\usepackage[dvips]{graphicx}
\usepackage{cite}
\usepackage{enumerate}
\usepackage{color}

\ifCLASSINFOpdf
\else
\fi
\begin{document}
\title{Advanced turning maneuver of a many-legged robot using pitchfork bifurcation}

\author{Shinya~Aoi$^1$, Ryoe~Tomatsu$^1$, Yuki~Yabuuchi$^1$, Daiki~Morozumi$^1$, Kota~Okamoto$^1$, Soichiro~Fujiki$^2$, Kei~Senda$^1$, and~Kazuo~Tsuchiya$^1$\\% <-this % stops a space
$^1$ Dept. of Aeronautics and Astronautics, Graduate School of Engineering, Kyoto University, Kyoto daigaku-Katsura, Nishikyo-ku, Kyoto 615-8540, Japan\\
$^2$ Dept. of Physiology and Biological Information, School of Medicine, Dokkyo Medical University, 880 Kita-Kobayashi, Mibu-machi, Shimotsuga-gun Tochigi 321-0293, Japan
%\thanks{S.~Aoi, R.~Tomatsu, Y.~Yabuuchi, D.~Morozumi, K.~Okamoto, K.~Senda, and K.~Tsuchiya are with the Dept. of Aeronautics and Astronautics, Graduate School of Engineering, Kyoto University, Kyoto daigaku-Katsura, Nishikyo-ku, Kyoto 615-8540, Japan e-mail: shinya\_aoi@kuaero.kyoto-u.ac.jp.}% <-this % stops a space
%\thanks{S.~Fujiki is with the Dept. of Physiology and Biological Information, School of Medicine, Dokkyo Medical University, 880 Kita-Kobayashi, Mibu-machi, Shimotsuga-gun Tochigi 321-0293, Japan.}% <-this % stops a space
}

\maketitle

\begin{abstract}
%多脚ロボットは多様な環境の踏破性に優れ、様々な場所での活躍が期待されている。
%しかしながら、多くの自由度の制御の困難さや地面に拘束される多くの脚が障害になり、機動的な運動の実現は未だ難しい問題である。
%従来の制御法では、目標とする運動を詳細に計画し、それを安定化するように制御することを目指しているが、ロボットや環境が複雑になると歩行性能はすぐに落ちてしまう。
%そのような精密な制御よりも、少ない情報に基づいたシンプルな制御が望ましく、そのためには環境との相互作用も込めた力学特性を利用することが重要である。
%そこで本研究では、柔軟な体軸を持った多脚ロボットを対象に、直線歩行の不安定性に着目した。
%具体的には、体軸柔軟性を変えることでピッチフォーク分岐を介して直線歩行が不安定化することを見出した。
%更には、不安定化した後に曲線歩行に移行することを見出した。
%この分岐特性を利用したシンプルな制御系を構築し、優れた旋回機動性を示すことを実証した。
%本制御系は、安定化を重視する従来の制御法とは大きく異なり、不安定性を積極的に利用し、更には分岐特性までも利用した特徴ある制御系であり、内在する力学特性を利用した多脚ロボットの機動性向上の優れた設計指針を与える。
Legged robots have excellent terrestrial mobility for traversing diverse environments and thus have the potential to be deployed in a wide variety of scenarios.
However, they are susceptible to falling and leg malfunction during locomotion.
Although the use of a large number of legs can overcome these problems, it makes the body long and leads to many legs being constrained to contact with the ground to support the long body, which impedes maneuverability.
To improve the locomotion maneuverability of such robots, the present study focuses on dynamic instability, which induces rapid and large movement changes, and uses a 12-legged robot with a flexible body axis.
Our previous work found that the straight walk of the robot becomes unstable through Hopf bifurcation when the body axis flexibility is changed, which induces body undulations.
Furthermore, we developed a simple controller based on the Hopf bifurcation and showed that the instability facilitates the turning of the robot.
In this study, we newly found that the straight walk becomes unstable through pitchfork bifurcation when the body-axis flexibility is changed in a way different from that in our previous work.
In addition, the pitchfork bifurcation induces a transition into a curved walk, whose curvature can be controlled by the body-axis flexibility.
We developed a simple controller based on the pitchfork-bifurcation characteristics and demonstrated that the robot can perform a turning maneuver superior to that with the previous controller.
This study provides a novel design principle for maneuverable locomotion of many-legged robots using intrinsic dynamic properties.
\end{abstract}


% \fi
\IEEEpeerreviewmaketitle

\section{Introduction}

%様々な生物に見られるように、脚歩行は、優れた不整地踏破性など陸上移動に有効な手段である。
%それゆえ、探索やレスキューミッション、危険区域での作業や探査、惑星探査など様々な場所での活躍が期待されている。
%更には、近年、動物のような俊敏な歩行を実現する脚ロボットも開発されている。
%しかしながら、これらの脚ロボットのほとんどは4本の脚しか持たず、多くてもせいぜい6本であり、機構系や電気系の故障を引き起こしやすい転倒からは逃れられず、一旦転倒すると回復することも難しい。
%更には、例え一本でも脚が故障すると、歩行機能は大幅に低下してしまう。
%ムカデのように多くの脚を用いると、そのような転倒や故障に耐性ができる。

\IEEEPARstart{L}{egged} locomotion, such as that of animals, allows excellent terrestrial mobility for traversing diverse environments.
Legged robots thus have potential to be deployed in a wide variety of scenarios, such as search and rescue~\cite{bib_hoffman1, bib_ning1}, hazardous environment operation and exploration~\cite{bib_byrd1, bib_toshiba1}, and planetary exploration~\cite{bib_arm1, bib_wilcox1}.
Various legged robots with agile animal-like locomotion have recently been developed~\cite{bib_aguilar1, bib_aoi3, bib_hwangbo1, bib_ijspeert1, bib_ijspeert2, bib_li2, bib_mastalli1, bib_owaki1, bib_park1, bib_raibert1, bib_steingrube1}.
However, most of these robots have four legs and falling, which may result in the breakdown of mechanical and electrical components and from which it is difficult to recover, is inevitable during locomotion.
Furthermore, damage to even one leg greatly degrades their locomotive performance~\cite{bib_cully1}.
The use of a large number of legs prevents falling and allows a certain level of leg malfunction to be tolerated~\cite{bib_kano1, bib_miguel1}.

%多くの脚の利用には有利な点があるものの、身体が長くなり、自由度の多さに起因する運動計画や制御の難しさや、環境との力学的な相互作用の取り扱いが難しいという問題がある。
%特に、接地する多くの足が地面に拘束されているため、それが機動的な歩行の障害となってしまう。
%ヒトや四足動物では旋回に身体を倒すなどの戦略が用いられるが、そのような拘束の中で、機動性のある歩行を実現するメカニズムは生物学的にも工学的にも不明確であり、多くの足を用いて素早い歩行をするロボットを作ることは未だとても難しく挑戦的である。

Although the use of a large number legs has advantages for legged robots, it makes the body long and increases the difficulty of motion planning and control due to the many intrinsic degrees of freedom and complex interaction with the environment.
In particular, many legs are physically constrained to be in contact with the ground to support the long body, which can impede maneuverability.
While humans and quadrupeds lean their bodies to enhance turning maneuvers, the underlying mechanism of agile locomotion using a large number of legs remains unclear from biological and engineering viewpoints~\cite{bib_full1}.
Thus, maneuverable locomotion for robots with a large number of legs remains challenging.

%機動性だけでなく、ロボットが自律的に歩き回るためには、状況に応じて目的地を変え、目的地に応じて適切な運動を作り出す必要がある。
%従来の制御では、例えば体軸をどのように曲げるか、どこに足を着くか、どの順で足を動かすかなど、全ての自由度の運動を入念に計画し、それらが精確に且つ安定に実現されるように制御する。
%しかしながら、そのような方法では、計算コストが非常にかかり、更にはエネルギー効率も悪い。

Conventional controllers precisely plan the motion of all degrees of freedom of the robot (e.g., how the long body is bent, where each foot touches the ground, and in what order the legs move) and control the robot to stabilize the desired motion.
However, this approach has huge computational and energy costs, making it inefficient.
To design a simple and efficient controller with high locomotor performance, the fundamental dynamic principles embedded in the robot dynamics, including the interaction with the environment, should be fully utilized~\cite{bib_aguilar1, bib_collins1, bib_li1, bib_li2}.

%本研究では、このような困難さや限界を克服して多脚ロボットの自律的で機動的な歩行を形成するために、運動を大きく変える駆動力となり得る動力学的不安定性に着目した斬新な制御法を提案する。
%具体的には、まず体軸を柔軟にした。
%多くの接地脚は機動性の障害になる可能性はあるものの、体軸柔軟性を変えると、ホップ分岐を介して直線歩行の不安定化を引き起こし、蛇行を生じさせることを先行研究で示している。
%分岐とはパラメータに応じて系の性質が変わり、その中でもホップ分岐とは、系の安定性の変化により平衡点から周期解を生み出す現象のこと
%直線歩行の不安定化により、容易で迅速に進行方向を変えることができ、機動性の向上に結びつく。

For maneuverable locomotion of many-legged robots that overcomes the above difficulties and the limitations of conventional approaches, the present study focuses on dynamic instability, which induces rapid and large movement changes, and uses a 12-legged robot whose body axis is flexible.
Our previous work~\cite{bib_aoi1} showed that although many ground contact legs can impede maneuverability, they induce straight walk instability and body undulations through Hopf bifurcation when the body-axis flexibility is changed; bifurcation qualitatively changes a dynamical system by changing a parameter, and Hopf bifurcation changes the stability of an equilibrium point and creates a limit cycle~\cite{bib_strogatz1}.
Stability refers to the capability to resist and recover from disturbances; straight walk instability is thus expected to allow the robot to easily change walking direction.
Therefore, we developed a simple controller based on the straight walk instability induced by the Hopf bifurcation to change walking direction without precise motion planning and control, and demonstrated that the straight walk instability facilitates the turning of the robot~\cite{bib_aoi2}.

%そこで本研究では、体軸剛性をこれまでとは違った形で変えることで実際にピッチフォーク分岐が生じさせることができ、新たな解は曲線形状を取り、その形状を制御できることを示す。
%その結果、従来に比べて、より機動性の向上に貢献することを示す。
%この制御法は、目標軌道の安定性を重視した従来のものとは違い、環境と相互作用するロボットに内在する不安定性を利用したものあり、今回更に分岐によって獲得した形状も利用することでより高度な特性を獲得している。
%この成果は、機動性の高い多脚歩行ロボットを構成する、力学を利用したシンプルな制御法の新たな設計原理に繋がると期待される。

In the present study, we show that the pitchfork bifurcation of the straight walk is caused by changes in the body-axis flexibility in a way different from that in previous work~\cite{bib_aoi1, bib_aoi2}; pitchfork bifurcation changes the stability of an equilibrium point and creates two equilibrium points~\cite{bib_strogatz1}.
The pitchfork bifurcation not only destabilizes straight walking, but also causes curved walking, where the flexible body axis forms a curved shape.
Furthermore, we found that the curvature of curved walking can be controlled by the body-axis flexibility.
We developed a simple control strategy based on the pitchfork-bifurcation characteristics, which improved the turning maneuver compared to that achieved using Hopf bifurcation.
This study also provides a design principle for a simple and efficient control scheme to create maneuverable locomotion for various robots using intrinsic dynamic properties.

\section{Robot}

%ロボットとして、先行研究で開発し、改良されたものを用いた。
%全長と全重は135cmと8.5kgである。
%図のように、6つの体節モジュールからなる。
%各モジュールは一つの胴体と1組の脚を持ち、長さは等しい。
%脚はピッチ関節でつながれた2リンクを持つが、先頭モジュールの脚は進行方向を変えるため更にヨー関節でつながれたリンクを持つ。
%脚関節は、エンコーダーの搭載されたモータで制御される。
%それぞれの体節は、回転バネとポテンショメータの搭載されたヨー関節でつながれている。
%真っ直ぐに並んだ時をゼロ度とする。
%分岐特性や旋回性能を調べるためにヨー関節には様々なばね定数を用いた。
%特に、ホップ分岐では、全てのヨー関節で同じバネ定数を用い、先行研究と同様次の7つを比較した。
%それに対して、ピッチフォーク分岐では、ヨー関節2-5で同じバネ定数を用い、ヨー関節1では次の7つを用いた。
%先頭は、旋回においてターゲットの相対位置を得るための測域センサが搭載されている。
%ロボットの電力は外部から供給し，外部のコンピュータ(RT-Linux)を用いて制御し、コマンドは2ms毎に受ける。
%歩行中脚先があまり滑らないように、ビニール製のフロアマットの敷かれた木製の床の上を歩く。
%ケーブルはロボットの歩行に影響がないように、つり上げられている。

\begin{figure}[t]
\begin{center}
	\includegraphics[width=78mm]{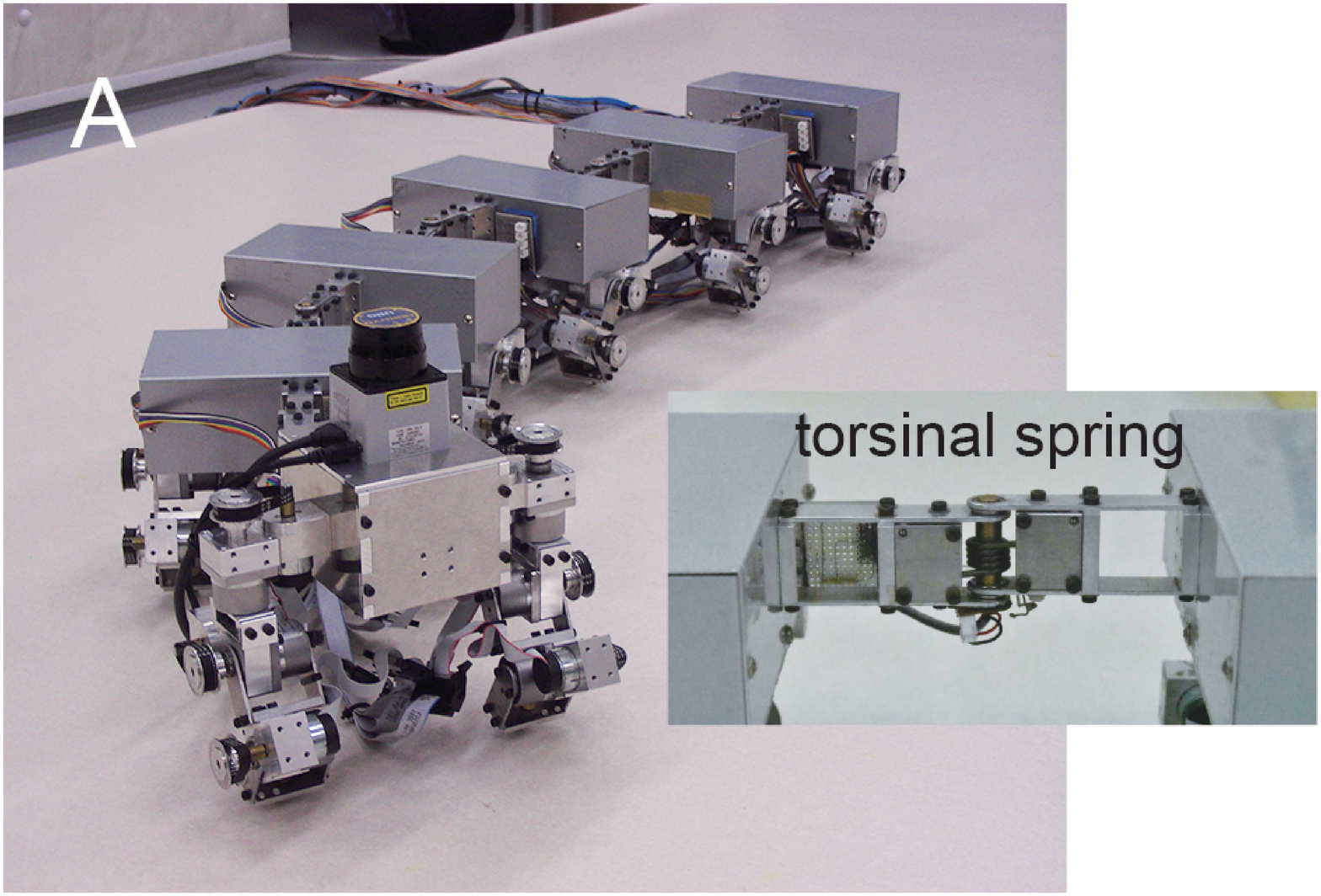}\vspace{3mm}
	\includegraphics[width=85mm]{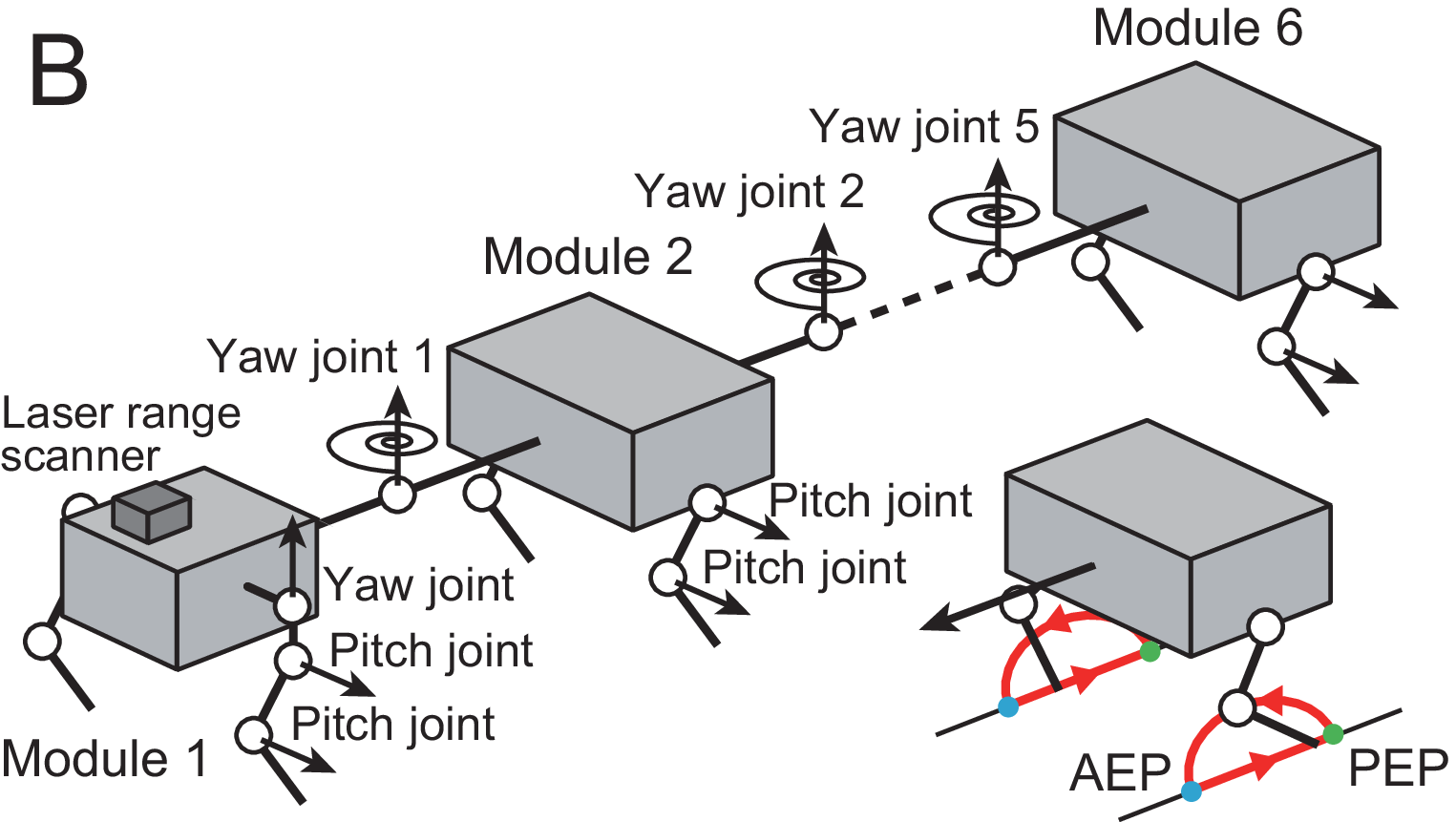}\vspace{0mm}

\caption{({\bf A}) Photograph and ({\bf B}) schematic model of many-legged robot. The robot consists of six modules, each of which has one body segment and one pair of legs. The legs are controlled by two pitch joints so that the leg tips follow a periodic trajectory, including the anterior extreme position (AEP) and the posterior extreme position (PEP). Body segments are passively connected by yaw joints with installed torsional springs. The legs in the first module have additional yaw joints to change the walking direction. A laser range scanner is installed on the first module to find a position relative to a target.
}
\label{robot}
\end{center}
\end{figure}

We used the many-legged robot developed in~\cite{bib_aoi1} and improved in~\cite{bib_aoi2}.
The total length and mass are 135~cm and 8.5~kg, respectively.
The robot consists of six body segment modules (modules 1--6), as shown in Fig.~\ref{robot}.
Each module is composed of a single body and one pair of legs and has the same length.
The body segments are passively connected by yaw joints (yaw joints 1--5) in which torsional springs (the spring constant is $k_i$ ($i=1,\dots,5$)) and potentiometers are installed at the axes.
The yaw joint angles are zero when the body segments are aligned.
Each leg has two links connected by pitch joints.
The legs in the first module (module~1) have an additional link connected by a yaw joint to supplement the control of the walking direction during turning tasks.
Each leg joint is manipulated by an encoder-equipped motor.
The first module has a laser range scanner with a viewing angle of 240~deg (Hokuyo, URG-04LX) to find the relative position of a target for turning.
The robot was controlled by an external host computer (Intel Pentium 4 2.8~GHz, RT-Linux) with 2-ms intervals, and walked on a flat wooden floor with a vinyl floor mat to suppress slipping.
The computer control signals and electric power were provided via external cables, which were kept slack and suspended to avoid influencing the locomotor behavior of the robot.

%\subsection{Controller and experiments}

%\subsubsection{Leg control for straight walk}

%ロボットの直線歩行を生成するために，脚のピッチ関節に取り付けられているモータを用いて，脚先が胴体に対して周期的な目標軌道を実現するように制御する(図~\ref{robot}{\bf b})．
%この目標軌道は，脚先が胴体に対して接地点(AEP)と離脱点(PEP)を結ぶ半楕円軌道(遊脚相)と直線軌道(支持脚相)を描くように設計されており，支持脚相では脚先は胴体に対して一定の速度で動く．
%すなわち，脚先を胴体に対して進行方向に対して常に平行に動かし，その方向に地面から推進力を受けることで歩行を生成する．
%ただし，それぞれのモジュールにおける左右の脚はそれぞれ逆位相で動かし，前後のモジュールにおける同側の脚は位相差$2\pi/3$で動かす．
%先頭モジュールにおいて脚先の軌道が胴体と平衡になるように脚のヨージョイントを固定しておく限り，体節のヨー関節にバネがあり、脚先は胴体に平衡に動かしているので、ロボットは進行方向に一定の速度で真っ直ぐ進むことが期待される．

To make the robot walk in a straight line, we controlled the legs using the two pitch joints of each leg to follow the desired movement, which consists of two parts, namely half of an elliptical curve that starts from the posterior extreme position (PEP) and ends at the anterior extreme position (AEP), and a straight line from the AEP to the PEP (Fig.~\ref{robot}{\bf B}).
In the straight line, the leg tips moved from the AEP to the PEP at a constant speed parallel to the body.
We set the duration of the half elliptical curve to $0.29$~s, that of the straight line to $0.31$~s, and the distance between the AEP and the PEP in each leg to $3$~cm.
The left and right legs in each module moved in antiphase, and the relative phase between the ipsilateral legs on adjacent modules was set to $2\pi/3$~rad.
When the leg yaw joint angles of the first module were fixed so that the leg tip trajectories were parallel to the body, the robot was expected to walk in a straight line while keeping the body segments parallel to each other because torsional springs were installed on the body-segment yaw joints and all support-leg tips moved parallel to the body segments at an identical speed.

\section{Pitchfork bifurcation of straight walk}

\subsection{Experimental results}
\label{sec_straight_exp}

%先行研究では、全てのヨー関節で同じ剛性を用い、それらを一様に変えるとホップ分岐を介して直線歩行が不安定化し、蛇行に至ることを確認していた。
%この分岐は、シンプルな数理モデルを用いて、周期係数系を持つ線形微分方程式の解の安定性を調べるフロケ解析を用いて検証していた。
%本研究では、ヨー関節2から5までの剛性を一定として、ヨー関節1のみ様々に変化させた。
%ただし、初期条件としては、全てのモジュールが真っ直ぐ並んだ状態から歩行を開始し、先頭の脚ヨージョイントは固定しておいた。

Our previous work~\cite{bib_aoi1, bib_aoi2} revealed that when we used torsional springs with the same spring constant for all body-segment yaw joints (yaw joints 1--5) and changed the spring constant uniformly among the joints, the straight walk became unstable through Hopf bifurcation, which induced body undulations.
This bifurcation was verified by a Floquet analysis, which investigates the stability of solutions of linear differential equations with periodic coefficients, with a simple physical model.
In this study, we performed robot experiments of walking in a straight line, where we changed the body-axis flexibility in a way that was different from that in our previous work.
Specifically, we used the same torsional spring for yaw joints 2--5 with $k_i=41$~Nmm/deg ($i=2,\dots,5$) and various torsional springs for yaw joint 1 with $k_1=15$, $17$, $21$, $28$, $41$, and $75$~Nmm/deg.
We set all the body segments parallel to each other as the initial conditions.
The leg yaw joints in the first module were fixed during the experiments and we did not attempt to control the walking direction, that is, the walking direction was an open loop.

\begin{figure*}[t]
\begin{center}
	\includegraphics[width=150mm]{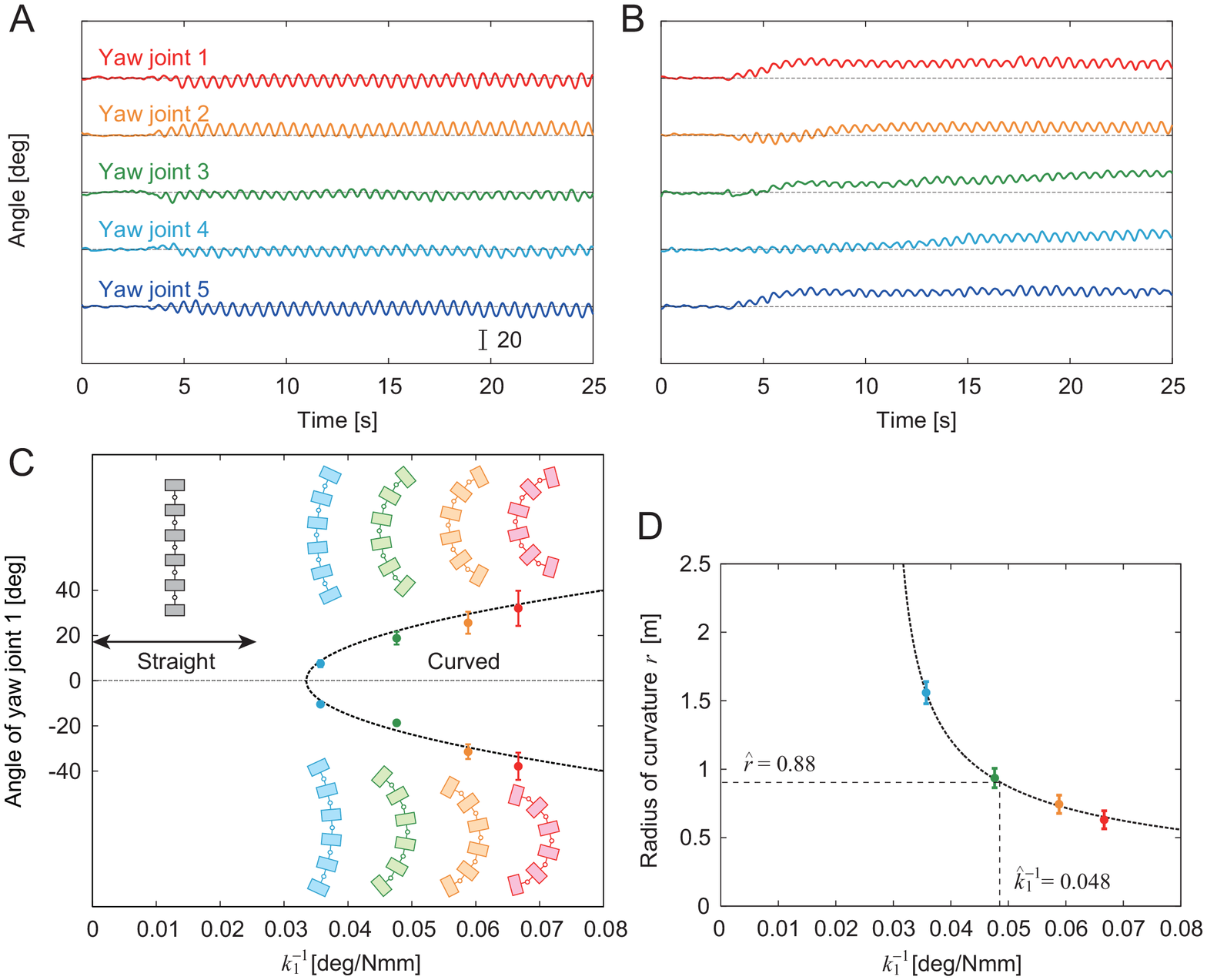}\vspace{3mm}
	\includegraphics[width=125mm]{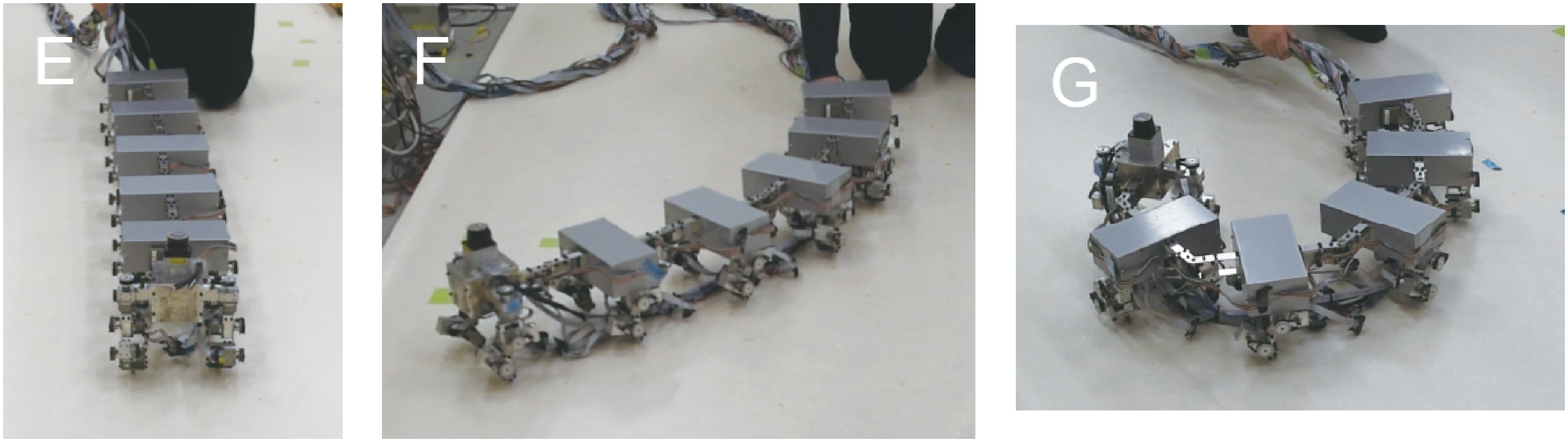}\vspace{0mm}

\caption{Characteristics of curved walk for $k_1$ values below the threshold value. Yaw joint angles for ({\bf A}) straight walk with $k_1=41$~Nmm/deg and ({\bf B}) curved walk with $k_1=15$~Nmm/deg (see Movie~1). ({\bf C}) Average angle of yaw joint~1 during curved walk for $1/k_1$ that indicates pitchfork bifurcation. The data points and error bars correspond to the means and standard errors, respectively, of the results of five experiments. ({\bf D}) Radius of curvature of body axis for $1/k_1$. The data points and error bars correspond to the means and standard errors, respectively, of the results of 10 experiments. Photographs of ({\bf E}) straight walk with $k_1>\hat{k}_1$, ({\bf F}) curved walk with small curvature with $k_1\sim\hat{k}_1$, and ({\bf G}) curved walk with large curvature with $k_1<\hat{k}_1$.}
\label{curved}
\end{center}
\end{figure*}

%実験の結果、ヨー関節1の剛性が大きい場合には、予想通り直線歩行が継続された（動画）。
%しかしながら、その剛性がある大きさを下回ると、体軸が円弧を描く曲線歩行が見られた（動画）。
%具体的には、次のバネ定数では曲線歩行を生じ、それ以外では生じず、曲線歩行では右に曲がる場合もあれば、左に曲がる場合もあった。
%図に、$k_1^{-1}$に対するヨー関節？の曲線歩行中の？秒間の平均角度を示す（他のヨー関節は付録）。
%足先の滑りのために関節毎の大きさが少し違っているが、体軸はほぼ曲線の形状を呈していることを示しており、
%振幅は$k_1^{-1}$に応じて変化しており、これらの結果は$k_1$に応じてピッチフォーク分岐が生じていることを示唆している。
%シンプルな数理モデルを用いたフロケ解析からも、ピッチフォーク分岐を示すことが示される（付録）。
%これらのデータは$k_1^{-1}$の1/2乗関数でフィッティングし、$k_1^{-1}=$？あたりに分岐点があることがわかる。

\begin{figure*}[t]
\begin{center}
	\includegraphics[width=150mm]{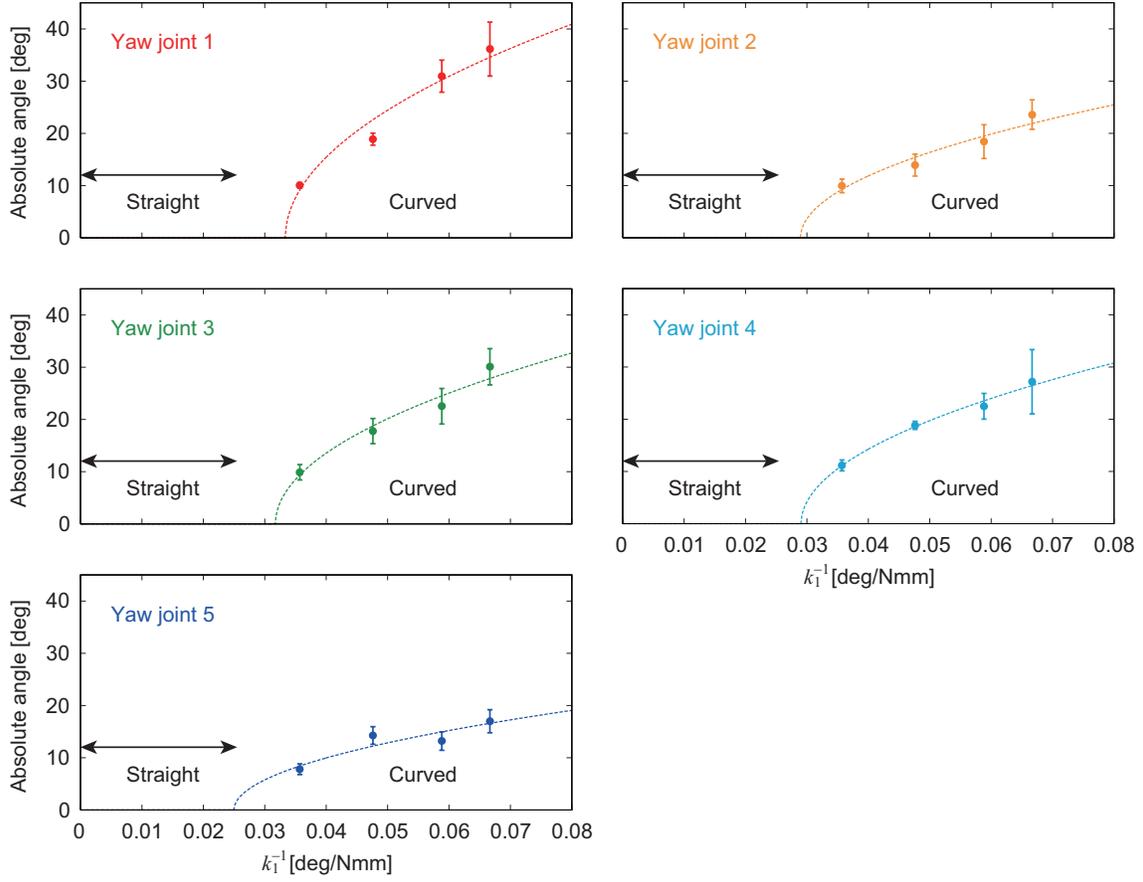}\vspace{0mm}

\caption{Average absolute angles of body-segment yaw joints during curved walk for $1/k_1$. The data points and error bars correspond to the means and standard errors, respectively, of the results of 10 experiments.}
\label{absolute}
\end{center}
\end{figure*}

When we used large spring constants for $k_1$, the robot kept walking in a straight line as expected, and the body segments were aligned, with all body-segment yaw joint angles being almost zero (Figs.~\ref{curved}{\bf A} and {\bf E}, see Movie~1).
However, when $k_1$ was set to below a threshold value, the body-segment yaw joints showed non-zero angles with the same sign; that is, the body axis was curved and the robot walked in a curved line (Figs.~\ref{curved}{\bf B}, {\bf F}, and {\bf G}, see Movie~1).
Specifically, the robot walked in a curved line for $k_1=15$, $17$, $21$, and $28$~Nmm/deg, but not for $k_1=41$ and $75$~Nmm/deg.
Furthermore, both left- and right-curved walking appeared.
Figure~\ref{curved}{\bf C} shows the angles of yaw joint~1 for $1/k_1$ averaged over 5~s during a curved walk (the angles for the other body-segment yaw joints are shown in Fig.~\ref{absolute}).
Although the angles slightly differ among yaw joints 1--5 partly due to feet slippage, these data indicate that the body axis shows a curved shape.
Furthermore, while the fluctuation among the trials increases with $1/k_1$, the magnitude of these angles increases with $1/k_1$.
These results suggest that the presence of pitchfork bifurcation depends on $k_1$.
These angle data were fitted by the square root of $1/k_1$~\cite{bib_strogatz1}.
The bifurcation point was estimated to be $k_1=34$~Nmm/deg ($1/k_1=0.030$~deg/Nmm).

%図におけるばね定数に対するヨー関節角度の変化は、ピッチフォーク分岐に従って曲線歩行を形成する体軸の曲率が変化していることを意味している。
%図に、という計算の下、バネ剛性に対する体軸の曲率を示す。
%体軸剛性に従って、曲線歩行を形成する旋回半径が変化していることがわかる。

The dependence of the body-segment yaw joint angles on $1/k_1$ (Fig.~\ref{curved}{\bf C} and Fig.~\ref{absolute}) indicates the change of the curved shape of the body axis for the curved walk.
Figure~\ref{curved}{\bf D} shows the radius of curvature $r$ of the body axis for $1/k_1$ calculated as $r=5L/\sum_{i=1}^{5}|\theta_i|$, where $\theta_i$ is the angle of yaw joint $i$ ($i=1,\dots,5$) and $L$ is the length of the body segments.
This figure shows that we can control the curvature of the body axis to perform a curved walk by adjusting $k_1$ through pitchfork bifurcation.

%このピッチフォーク分岐は$k_1$以外の他のパラメータにも依存する。
%これを調べるために、ヨー関節2-5の剛性と歩行速度、そして脚運動の前後位相差を用いた。
%具体的には、剛性を28にする、もしくはAEPとPEPの距離を2cmにする、もしくは位相差を$5\pi/6$にした。
%図に得られた曲率を示す。

\begin{figure}[t]
\begin{center}
	\includegraphics[width=80mm]{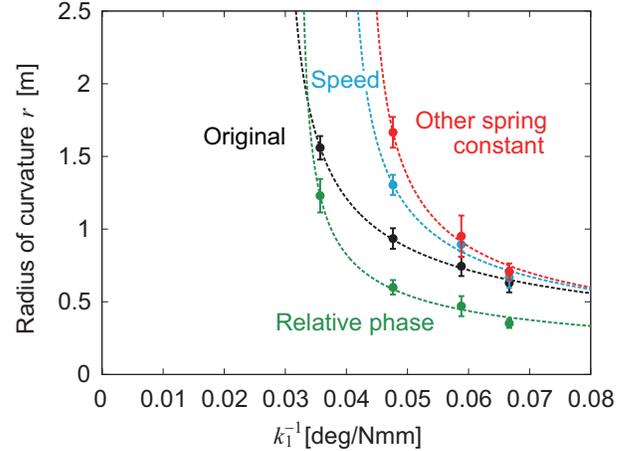}\vspace{0mm}

\caption{Change in radius of curvature by different values for the spring constant of yaw joints 2--5, gait speed, and relative phase between the ipsilateral legs on adjacent modules. The data points and error bars correspond to the means and standard errors, respectively, of the results of five experiments.
}
\label{different}
\end{center}
\end{figure}

The pitchfork bifurcation also depends on parameters other than $k_1$.
To investigate this, we used different values for the spring constant of yaw joints 2--5, gait speed, and relative phase between the ipsilateral legs on adjacent modules.
Specifically, we changed $k_{2-5}$ from $41$ to $28$~Nmm/deg, the distance between the AEP and the PEP in each leg (Fig.~\ref{robot}{\bf B}) from $3$ to $1.8$~cm, or the relative phase from $2\pi/3$ to $\pi$, and performed the same experiments shown in Fig.~\ref{curved}.
As a result, these values also induced a curved walk below a critical value of $k_1$.
However, the estimated bifurcation point and radius of curvature of the body axis changed as shown in Fig.~\ref{different}.
Specifically, when the spring constant of yaw joints 2--5 and gait speed decreased, the estimated bifurcation point decreased from $k_1=34$ to 23 and 25~Nmm/deg ($1/k_1=0.030$ to 0.043 and 0.040~deg/Nmm), respectively.
However, the radius of curvature remained almost unchanged.
In contrast, while the relative phase did not change the bifurcation point as much ($k_1=31$~Nmm/deg, $1/k_1=0.032$~deg/Nmm), it achieved a smaller radius of curvature.

\subsection{Verification by Floquet analysis with simple physical model}

%実験の結果から、直線歩行のピッチフォーク分岐が見られることが推測された、
%先行研究と同様、シンプルな数理モデルを用いて理論的な視点からこの分岐の妥当性を検証した。
%モデルは歩行の本質を抜き出すようにもとの複雑なモデルから簡略化されている。
%特に、多脚ロボットは平面的に歩行するように設計されているので、モデルは2次元となっている。
%更に、脚の主な役割は地面からの床反力なので、慣性力は無視して、脚先の幾何学的な拘束力を用いている。
%具体的には、脚先が体節に対して目標通り動くものとして、AEPからPEPまで速度に比例する摩擦力を受けるものとしている。

\begin{figure}[t]
\begin{center}
	\includegraphics[width=75mm]{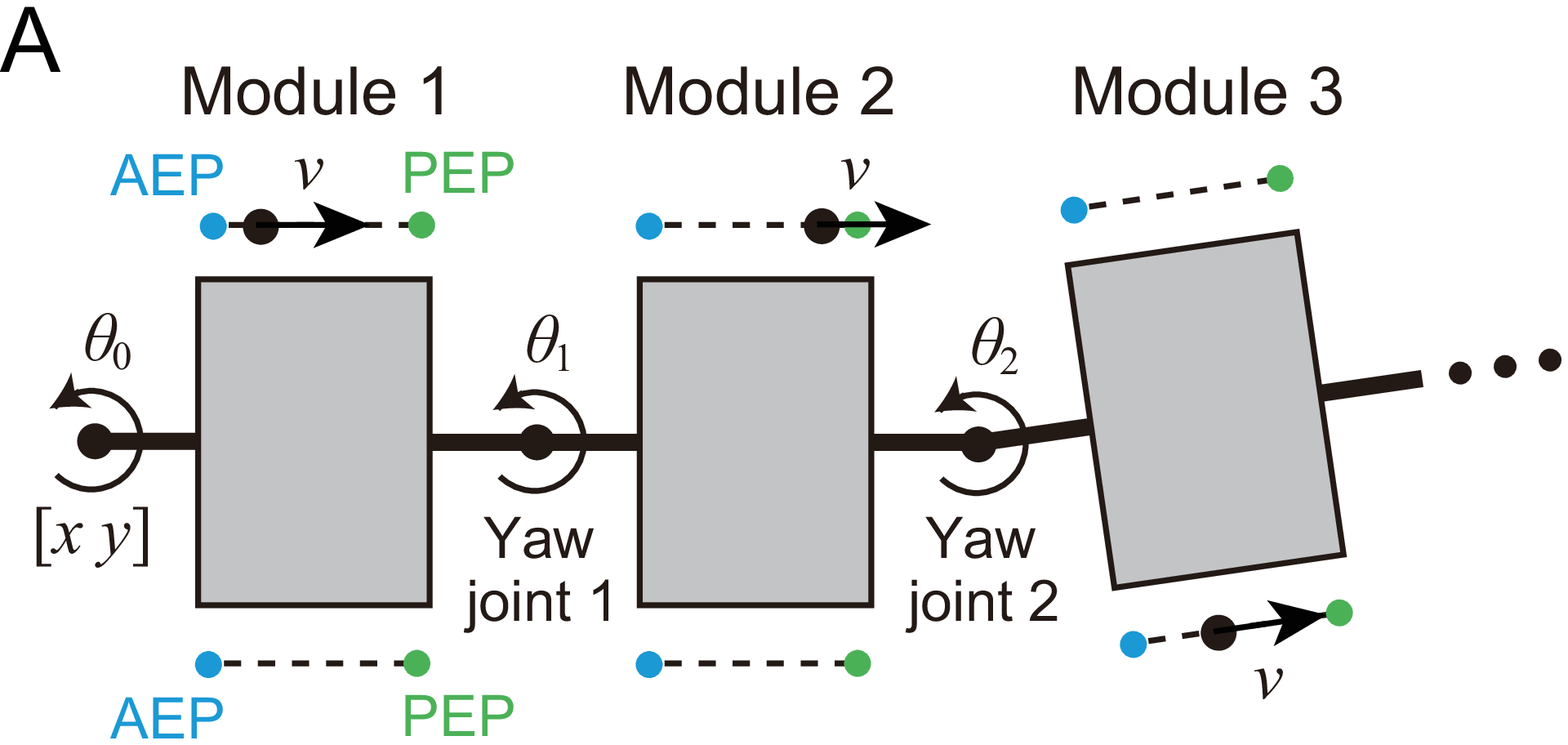}\vspace{1mm}
	\includegraphics[width=82mm]{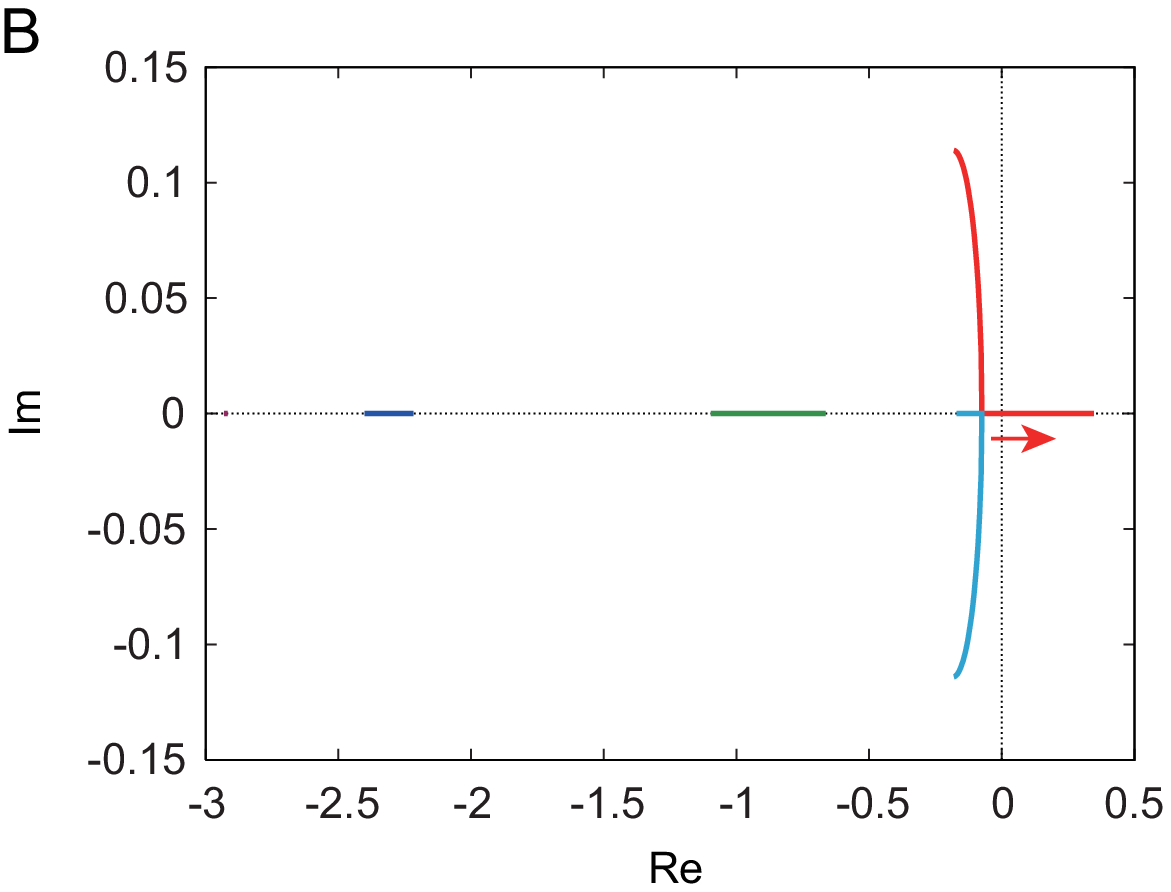}\vspace{1mm}
	\includegraphics[width=85mm]{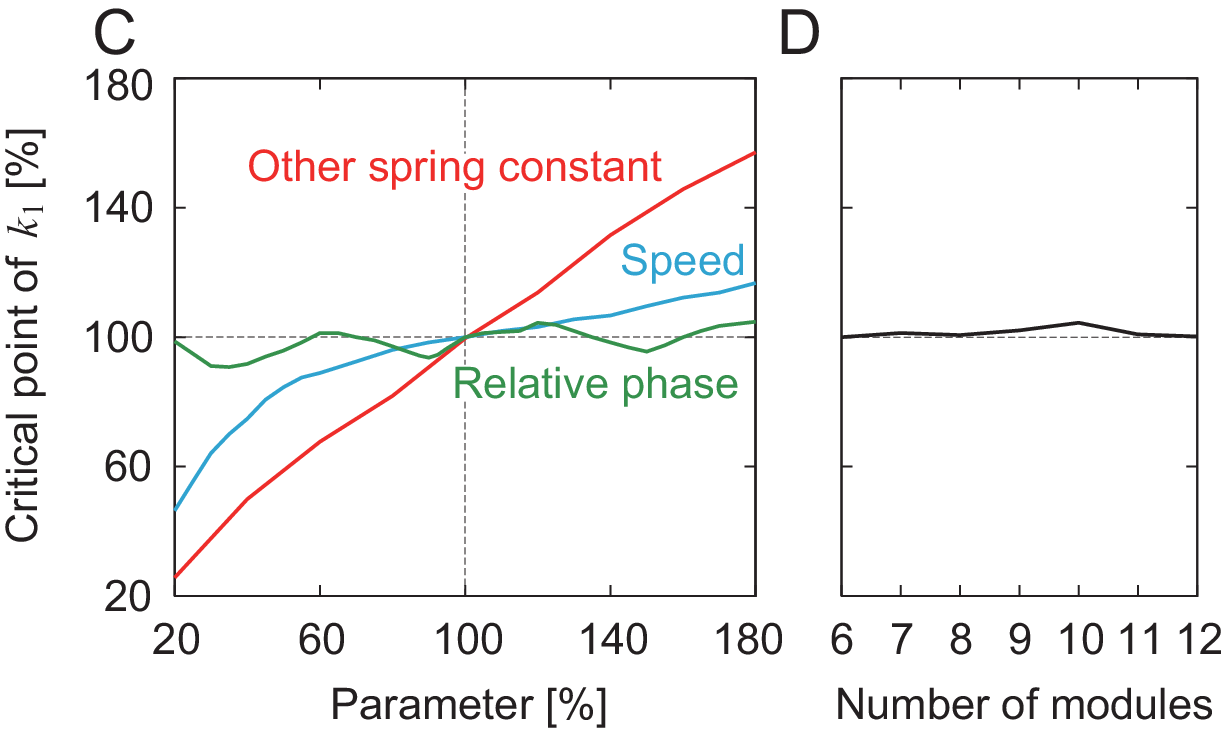}\vspace{0mm}

\caption{Floquet analysis using simple two-dimensional model. ({\bf A}) Simple model. ({\bf B}) Floquet exponents when $k_1$ was varied. Different colors represent different exponents. The exponent in red crosses into the right-half plane along the real axis, indicating pitchfork bifurcation. The change in critical value of $k_1$ by ({\bf C}) spring constant of yaw joints 2--5, gait speed, and relative phase between the ipsilateral legs on adjacent modules, and ({\bf D}) number of modules.}
\label{simple}
\end{center}
\end{figure}

The robot experiments suggested that the presence of pitchfork bifurcation in the straight walk depends on the spring constant $k_1$ (Figs.~\ref{curved}{\bf C}, \ref{absolute}, and \ref{different}).
We verified this bifurcation from a theoretical viewpoint using a Floquet analysis with a simple physical model, as done in our previous work~\cite{bib_aoi1}.
The model was simplified from the original high-dimensional mechanical model to extract the fundamentals of locomotion dynamics (Fig.~\ref{simple}{\bf A}).
In particular, the model was two-dimensional because the movements were designed to make the robot walk without up-and-down, roll, or pitch motions of the body segments.
Furthermore, because an important role of legs in locomotion is to receive reaction forces from the floor, we neglected the inertial force of the legs and instead modeled the reaction forces at the leg tips based on the geometric conditions.
Specifically, we assumed that the leg tips move relative to the body segments as designed and that the leg tips receive the friction forces during the straight line from the AEP to the PEP (Fig.~\ref{robot}{\bf B}), which are proportional to the velocities relative to the floor.

%このシンプルモデルの運動方程式を直線歩行のまわりで線形化し、複素平面上でのフロケ指数を調べた。
%先行研究では、全てのヨー関節の剛性を一様に小さくすると、一組のフロケ指数が左半平面から虚軸を交差して右半平面に移り、ホップ分岐を示した。

The equations of motion of the simple model can be expressed as
\begin{eqnarray}
K(q)\ddot{q}+h(q,\dot{q})=u(q,\dot{q})+\lambda(q,\dot{q},t) \label{EOM}
\end{eqnarray}
where $q=[x\,y\,\theta_0\,\cdots\,\theta_5]^{\scriptsize{\mbox{T}}}$, $[x\,y]$ and $\theta_0$ are the position and yaw angle of the first module, respectively; $K(q)$ is the inertia matrix; $h(q,\dot{q})$ is the nonlinear term; $u(q,\dot{q})$ is the torque term of the torsional springs; and $\lambda(q, \dot{q},t)$ is the reaction force term.
Because the leg tips move periodically relative to the body segments, the reaction force $\lambda$ becomes a function of time $t$.
The detailed description and derivation of the equations of motion (\ref{EOM}) are shown in our previous work~\cite{bib_aoi1}.
During the straight walk of the model, we can write $\hat{q}=[vt+x_0\,y_0\,0\,\cdots\,0]^{\scriptsize{\mbox{T}}}$ and $\dot{\hat{q}}=[v\,0\,0\,\cdots\,0]^{\scriptsize{\mbox{T}}}$, where $x_0$ and $y_0$ represent the initial position of the first module and $v$ is the velocity of the leg tips relative to the body segments.
For $z^{\scriptsize{\mbox{T}}}=[q^{\scriptsize{\mbox{T}}}\,\dot{q}^{\scriptsize{\mbox{T}}}]$, the linearization of the equations of motion (\ref{EOM}) for a straight walk using $z=\hat{z}+\delta z$ gives
\begin{eqnarray}
\delta \dot{z}=A(t)\delta z \label{Linear}
\end{eqnarray}
Because the movements of the leg tips are periodic with the gait cycle $\tau$, $A(t+\tau)=A(t)$ is satisfied.
The detailed description and derivation of the linearized equation (\ref{Linear}) are also shown in our previous work~\cite{bib_aoi1}.
The fundamental solution matrix $Z(t)$ of the linearized equation with periodic coefficients (\ref{Linear}) can be expressed as
\begin{eqnarray}
Z(t)=\Phi(t)e^{\Lambda t}
\end{eqnarray}
where $\Phi(t+\tau)=\Phi(t)$~\cite{bib_guckenheimer1}.
Because we can use an identity matrix for $Z(0)$ and $\Phi(0)$, the integration of (\ref{Linear}) from $t=0$ to $\tau$ yields
\begin{eqnarray}
Z(\tau)=\Phi(\tau)e^{\Lambda\tau}=\Phi(0)e^{\Lambda\tau}=e^{\Lambda\tau}
\end{eqnarray}
The Floquet exponents (eigenvalues of the constant matrix $\Lambda$) and corresponding eigenvectors explain the behavior of the model.
Specifically, when all real parts of the exponents are negative, the straight walk of the model is asymptotically stable.
In contrast, if any real part of the exponents is positive, the straight walk becomes unstable.
In our previous work~\cite{bib_aoi1}, when all the spring constants of yaw joints 1--5 of the simple model were decreased uniformly, one pair of the Floquet exponents crossed the imaginary axis from the left-half plane and entered the right-half plane above a critical value of the spring constant, which indicates an oscillatory destabilization of the straight walk to produce body undulations and implied Hopf bifurcation (strictly speaking, this corresponds to Neimark-Sacker bifurcation when considering the periodicity of the gait cycle~\cite{bib_kuznetsov1}).
%Furthermore, not only was the bifurcation point of the model comparable to that of the robot experiments, but also the ratio of the amplitudes and phases of the components of the destabilizing eigenvector were comparable to those of the yaw joint movements during the robot experiments.
Furthermore, the relative amplitude and phase between the components of the destabilizing eigenvector were comparable to those of the yaw joint movements during the robot experiments.
The Floquet analysis using a simple model is useful for verifying the bifurcation observed in the robot experiments.

%図に、ロボット実験と同様、ヨー関節２−５の剛性は固定のまま、ヨー関節１の剛性のみ小さくしたときの複素平面上でのフロケ指数を示す。
%幾つかのゼロ固有値を除いて、剛性が大きいときはフロケ指数は全て左半平面に存在する。
%しかしながら、剛性を下げていくと、一つのフロケ指数が実軸上を移動して、右半平面に至っている。
%すなわち、剛性の臨界値を越えて直線歩行が不安定化し、ピッチフォーク分岐が生じていることが示される。
%更に、この分岐点で不安定化した指数に対応する固有ベクトルは次のような要素をもつ。
%実験で見られたのと同じように、それぞれ同じ符号で、関節1のものが少し大きく、関節5のものが小さいことがわかる。
%更に、図に位相差やk2-5の値、更には体節数を変化させたときの、分岐点の変化を示す。
%線形化方程式に基づいているので、曲率までは評価できないが、これが実験における、ピッチフォーク分岐を介した直線歩行の不安定化と曲線歩行の生成の確認となる。

Figure~\ref{simple}{\bf B} shows the Floquet exponents of the simple model when $k_1$ was varied while the other spring constants $k_i$ ($i=2,\dots,5$) remained fixed, as done in the robot experiments.
Except for the zero exponents, all exponents lie in the left-half plane for large $k_1$.
However, with decreasing $k_1$, one exponent moves along the real axis and enters the right-half plane above $k_1=12$~Nmm/deg.
%, which is close to the bifurcation point estimated in the robot experiments in Section~\ref{sec_straight_exp}.
%Furthermore, the components of the destabilizing eigenvector for yaw joints 1--5 at the critical point were $0.31$, $0.16$, $0.21$, $0.21$, and $0.11$, respectively, showing the same sign.
Although this critical value is smaller than the bifurcation point estimated in the robot experiments in Section~\ref{sec_straight_exp}, the components of the destabilizing eigenvector for yaw joints 1--5 at the critical point were $0.64$, $0.36$, $0.48$, $0.46$, and $0.16$, respectively, showing the same sign.
This indicates that the straight walk is destabilized to produce a curved walk (positive disturbance induces right-curved walk and negative disturbance induces left-curved walk) and implies pitchfork bifurcation.
Furthermore, the ratio between the components of the destabilizing eigenvector is consistent with the ratio between the yaw joint angles during the curved walk of the robot experiments (Fig.~\ref{absolute}).
Although this analysis does not estimate the radius of curvature after the bifurcation due to the limitation of the linear analysis, these results verify the destabilization of the straight walk and the emergence of a curved walk through pitchfork bifurcation, as observed in the robot experiments.

%この分岐における他のパラメータの影響についても調べた。
%特に、実験と同様、ヨー関節2-5の剛性と歩行速度、そして脚運動の前後位相差を用いた。
%それに加えて、モジュール数も変えた。
%$k_1$の分岐点がどのように変化するかを調べた。
%図に結果を示す。
%ここでは、この4つの影響しか調べていないが、ロボットの重さや長さなど、他のパラメータの影響についても調べることができる。

We also investigated the parameter dependence of the pitchfork bifurcation.
Figure~\ref{simple}{\bf C} shows how the critical value of $k_1$ changes as the spring constant of yaw joints 2--5, gait speed, and relative phase between the ipsilateral legs on adjacent modules change in the same way as in the robot experiments in Fig.~\ref{different}.
In addition, Fig.~\ref{simple}{\bf D} shows the change in the critical value when the number of the body segments is increased.
When the spring constant of yaw joints 2--5 and gait speed decrease, the critical value decreases.
However, the relative phase does not change the critical value as much.
These results are consistent with those in the robot experiments (Fig.~\ref{different}).
The number of body segments also does not change the critical value very much.
Although we investigated the effects of only changes in the spring constant of yaw joints 2--5, gait speed, relative phase, and number of body segments, this Floquet analysis can investigate other parameters, such as the length and mass, as performed in our previous work~\cite{bib_aoi1} for Hopf bifurcation.

\section{Turning maneuverability}

\subsection{Turning strategy based on pitchfork bifurcation}

%先行研究と同様、ロボットの機動性を調べるために、向いている方向とは大きく異なる場所におかれたターゲットへと近づける急旋回タスクの実験を行った。
%図のように、任意の位置のターゲットに対して、曲線歩行で近づくことができる曲率半径$\hat{r}$が一意に存在する。
%図で得られたように、ピッチフォーク分岐により形成される曲線歩行の体軸の曲率半径は、体軸剛性に対して単調に変化するので、その曲率半径が$\hat{r}$を形成する体軸剛性を一意に決定することができる。
%すなわち、この剛性を選べば、ピッチフォーク分岐の特性を利用することで自然とターゲットに近づくことができるはずである。
%ただし、この方法は初期の位置関係に応じたフィードフォワード制御であり、更にはピッチフォーク分岐では左右どちらに旋回するかは初期のロボットの状態に依存するので、ターゲット追従は保証されない。
%そのため、先行研究で構成した先頭の測域センサに基づく先頭の脚のヨー関節の制御も併用した。
%このフィードバック制御を使うと、$k_1\ne\hat{k}_1$でもターゲットに近づけることができる。

\begin{figure*}[t]
\begin{center}
	\includegraphics[width=150mm]{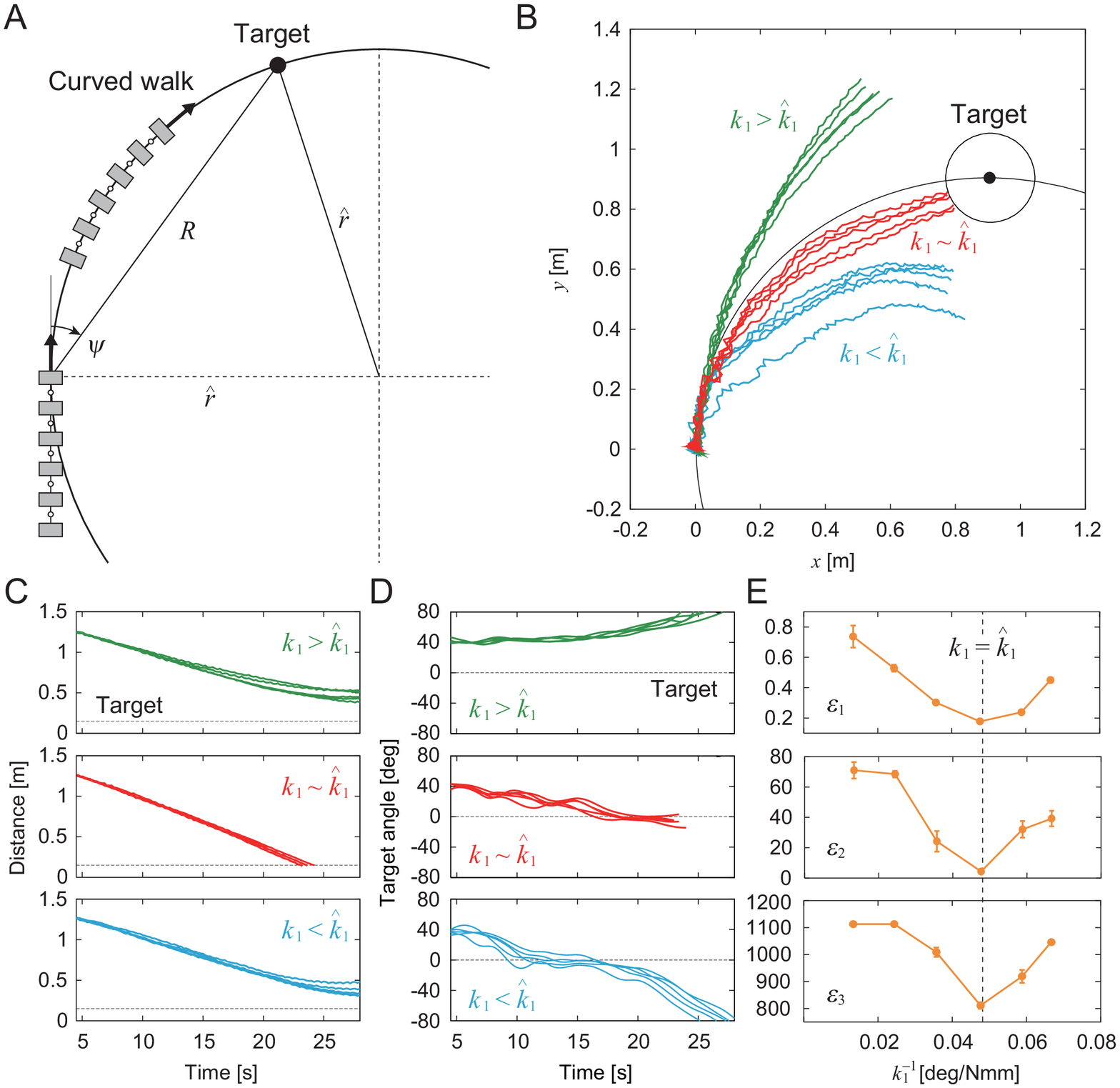}\vspace{2mm}
	\includegraphics[width=140mm]{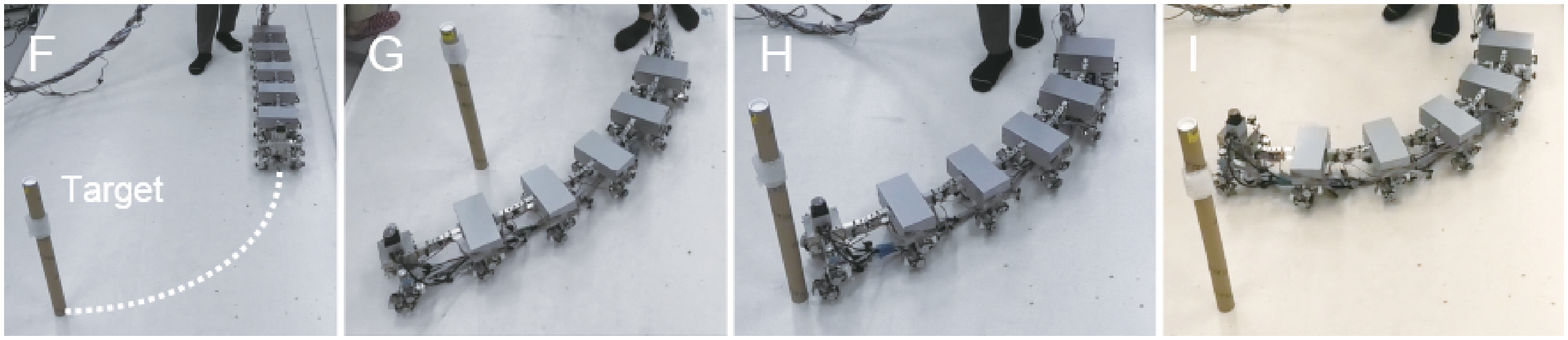}\vspace{-2mm}

\caption{Turning task. ({\bf A}) Radius of curvature $\hat{r}$ of curved walk with which the robot approaches a target (relative angle $\psi$ and distance $R$). ({\bf B}) Trajectory of the first module on the floor, ({\bf C}) target distance, and ({\bf D}) relative target angle for five experiments for three spring constants with $\psi=45^\circ$, $R=1.3$~m, $\hat{r}=0.88$~m, and $1/\hat{k}_1=0.048$~deg/Nmm (see Movie~2). ({\bf E}) Evaluation criteria $\varepsilon_1$, $\varepsilon_2$, and $\varepsilon_3$ for $1/k_1$. The data points and error bars correspond to the means and standard errors, respectively, of the results of five experiments. Photographs of ({\bf F}) initial conditions, ({\bf G}) unsuccessful approach with $k_1>\hat{k}_1$, ({\bf H}) successful approach with $k_1\sim\hat{k}_1$, and ({\bf I}) unsuccessful approach with $k_1<\hat{k}_1$.}\vspace{-4mm}
\label{turning}
\end{center}
\end{figure*}

To investigate the maneuverability of the robot achieved with the aid of pitchfork bifurcation, we focused on a turning task in which the robot approached a target located on the floor in a direction different from that to which the robot was oriented, as performed in our previous work~\cite{bib_aoi2}, which investigated the maneuverability with the aid of Hopf bifurcation.
For a target at any location (relative angle $\psi$ and distance $R$), there exists a unique radius of curvature $\hat{r}$ of the curved walk with which the robot will approach the target (Fig.~\ref{turning}{\bf A}).
Because the radius of curvature $r$ of the body axis induced by the pitchfork bifurcation monotonically decreases with $1/k_1$ (Fig.~\ref{curved}{\bf D}), $k_1=\hat{k}_1$ is uniquely determined so that $r=\hat{r}$.
This means that when we use $k_1=\hat{k}_1$, the robot spontaneously approaches the target due to the pitchfork bifurcation characteristics, which is an optimal strategy for turning.
However, this strategy is feedforward, depending on the initial conditions of the robot and target, that is, an open loop for the walking direction.
In particular, the direction in which the robot turns (left or right) depends on the initial robot conditions because of the pitchfork bifurcation characteristics, and thus this strategy does not guarantee the success of the turning tasks.
Therefore, we also used a supplementary turning controller developed in our previous work~\cite{bib_aoi2}, which uses the laser range scanner of the first module to measure the relative target angle and manipulates the leg yaw joints of the first module to approach the target based on the measured target angle (see Appendix~\ref{Append_controller}).
This supplementary controller is a closed loop for the walking direction and allows the robot to approach the target even when $k_1\ne\hat{k}_1$.

\subsection{Experimental results}

%初期条件として，ターゲットの相対角は$\theta_t=45^\circ$，距離は$r=1.3$~mとした。
%このとき、$\hat{r}=0.92$~m、$\hat{k}_1=?$~Nmm/degとなる。
%また、全ての体節間ヨージョイントは零とした状態から開始した。
%この旋回性能を調べるために、次のようなヨー関節1のばね定数に対して実験を行った。
%距離が15cm以内になると、旋回タスクは完了となる。
%図に、3つのバネ剛性に対する旋回タスクにおける先頭モジュールの位置の結果を示す。
%図に、3つのバネ剛性に対する旋回タスクにおけるターゲットに対するロボットの進行方向と距離の時間履歴を示す。
%$k_1\gg\hat{k}_1$では、素早く進行方向を変えることができず、軌道もオーバーシュートしてしまい、ターゲットには到達しなかった。
%$k_1\ll\hat{k}_1$では、素早く進行方向を変えることはできたものの、ピッチフォーク分岐による曲率半径が小さすぎるため、ターゲットの手前で曲がってしまい、到達しなかった。
%それに対して、$k_1\sim\hat{k}_1$では、ほぼピッチフォーク分岐による曲線歩行に沿うような旋回が実現した。

For the initial conditions, we used $\psi=45^\circ$ and $R=1.3$~m for the relative angle and distance between the first module and the target, respectively, which yielded $\hat{r}=0.88$~m and $\hat{k}_1=21$~Nmm/deg ($1/\hat{k}_1=0.048$~deg/Nmm), and set all body-segment yaw joint angles to zero (Fig.~\ref{turning}{\bf F}).
Figure~\ref{turning}{\bf B} shows the trajectory of the first module on the floor during the turning task for three torsional spring constants, namely $k_1=15$ ($<\hat{k}_1$), $21$ ($\sim\hat{k}_1$), and $41$~Nmm/deg ($>\hat{k}_1$).
Figures~\ref{turning}{\bf C} and {\bf D} show the time profiles of the target distance and relative target angle with respect to the walking direction, respectively, for these three spring constants.
When the distance was less than 0.15~m, we assumed that the robot reached the target and this task was successfully completed.
For $k_1=41$~Nmm/deg ($>\hat{k}_1$), the robot hardly changed walking direction and the first module trajectory bulged outward.
As a result, the robot could not reach the target (Fig.~\ref{turning}{\bf G}, see Movie~2).
For $k_1=15$~Nmm/deg ($<\hat{k}_1$), although the robot could quickly change walking direction, it moved in directions away from the target due to the small radius of curvature created by pitchfork bifurcation and could not reach the target (Fig.~\ref{turning}{\bf I}, see Movie~2).
In contrast, for $k_1=21$~Nmm/deg ($\sim\hat{k}_1$), the robot reached the target through the optimal curved walk generated by pitchfork bifurcation (Fig.~\ref{turning}{\bf H}, see Movie~2).

%これらバネ定数に依存した旋回性能を定量的に明らかにするために、2つの評価量を調べた。
%一つ目と二つ目の指標は、どれだけ早くターゲットに近づいたかを評価するためのターゲットの距離と相対角に関するものである。
%一つ目の評価量は、開始から時刻までのターゲットとの距離を積分したものであり、どれほど早くターゲットに近づくのかを評価している。
%二つ目の評価量は、開始から時刻までのターゲットとロボットの進行方向の差(絶対値)を積分したものであり、どれほど早くターゲットの方向を向けているのかを評価している。
%図にその結果を示す。
%それぞれ$1/\hat{k}_1$近傍で最小値を取っており、最適戦略において最も良い旋回性能が得られており、ピッチフォーク分岐により生成される曲線歩行を最大限利用していることがわかる。

To quantitatively clarify the turning performance dependence on $k_1$, we employed three evaluation criteria, namely $\varepsilon_1$, $\varepsilon_2$, and $\varepsilon_3$.
For the criterion $\varepsilon_1$, we used the distance of the target at 23~s (the earliest time at which the task is successfully completed) to evaluate how quickly and successfully the robot approached the target.
For the criterion $\varepsilon_2$, we used the absolute value of the relative target angle with respect to the walking direction at 23~s to evaluate how quickly and successfully the robot was oriented to the target.
For the criterion $\varepsilon_3$, we used the amount of the control input during the task from the supplementary turning control in the leg yaw joints of the first module (Appendix~\ref{Append_controller}) to evaluate how efficiently the robot performs turning.
Specifically, $\varepsilon_3=\int(\hat{\psi}_1^2+\hat{\psi}_2^2)dt$.
Figure~\ref{turning}{\bf E} shows the results for $1/k_1$.
All criteria showed minimum values around $k_1=\hat{k}_1$, which means that the turning strategy using pitchfork bifurcation achieved the best performance and that the robot made the best use of the curved walk induced by pitchfork bifurcation to complete the turning task.

%ピッチフォーク分岐を利用した提案する旋回制御則の有効性を検証するため、先の図とは異なる次の初期条件を用いて同様の実験を行った。
%図に剛性に対する評価指標を示す。
%それぞれ$1/\hat{k}_1$近傍で最小値を取っており、前の結果と同様、最適戦略において最も良い旋回性能が得られている。
%ターゲットに関する他の初期条件($\psi=40^\circ$、$R=1.5$~m)でも同様の結果が得られており、この提案手法の妥当性を示す。

\begin{figure}[t]
\begin{center}
	\includegraphics[width=88mm]{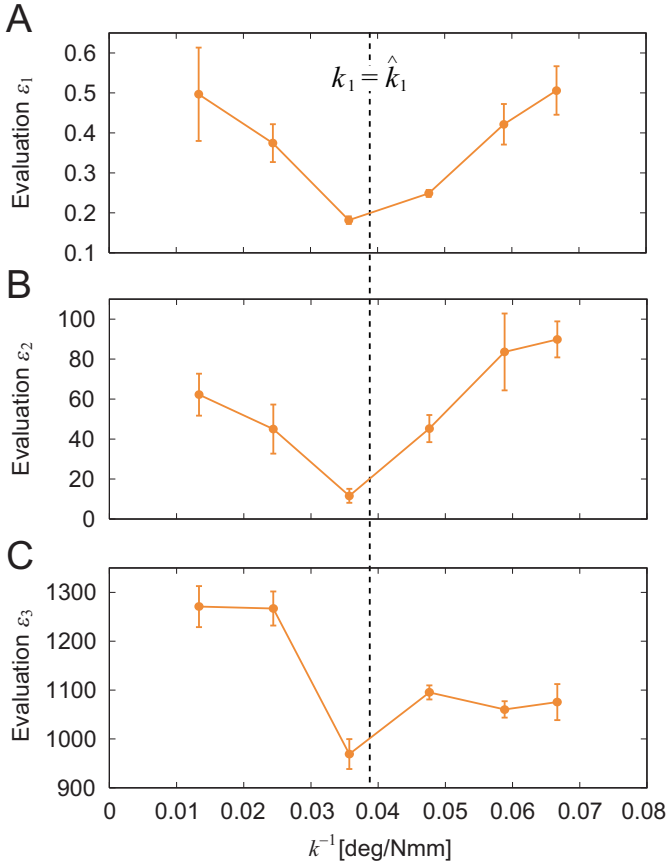}\vspace{0mm}

\caption{Evaluation criteria ({\bf A}) $\varepsilon_1$, ({\bf B}) $\varepsilon_2$, and ({\bf C}) $\varepsilon_3$ for $1/k_1$ for different conditions ($1/\hat{k}_1=0.039$~deg/Nmm). The data points and error bars correspond to the means and standard errors, respectively, of the results of five experiments.}
\label{another}
\end{center}
\end{figure}

To verify the performance of the proposed controller using pitchfork bifurcation, we additionally performed the same experiment as that in Fig.~\ref{turning}{\bf B} but using different initial conditions of the target, namely $\psi=40^\circ$ and $R=1.5$~m, which yielded $\hat{r}=1.2$~m and $\hat{k}_1=26$~Nmm/deg ($1/\hat{k}_1=0.039$~deg/Nmm).
Figures~\ref{another}{\bf A}, {\bf B}, and {\bf C} show the evaluation criteria $\varepsilon_1$, $\varepsilon_2$, and $\varepsilon_3$, respectively, for $1/k_1$.
All criteria show minimum values around $k_1=\hat{k}_1$, which means that the turning strategy using pitchfork bifurcation achieved the best performance, in the same way as shown in Fig.~\ref{turning}{\bf E}.
The results show similar trends, which verifies the performance of the proposed controller.

\subsection{Contribution of supplementary turning control}

%ピッチフォーク分岐によって形成された曲線歩行を利用することで、効率良く機動的にターゲットに近づく旋回ができることが示された。
%しかしながら、最初体軸を真っ直ぐな状態から開始すると、ピッチフォーク分岐のみでは曲線歩行に収束するにも図のように少し時間がかかる。
%更に、初期の直線歩行はピッチフォーク分岐の不安定解に対応するため、どちらに旋回するかはわからない（摂動が旋回方向を決め、摂動がなければ、直線歩行が持続される）。
%そのため、例え$k_1=21\sim\hat{k}_1$の最適なバネ剛性を用いたとしても、ピッチフォーク分岐の特性がターゲットへの到達を補償するわけではない。
%これを補償するために、補助的な旋回制御を用いた。

As demonstrated above, the robot achieved maneuverable and efficient turning to approach a target using the curved walk induced by the pitchfork bifurcation.
However, when the robot starts the approach with the body axis straight, it takes some time for the convergence to a curved walk, as shown in Fig.~\ref{curved}{\bf B}.
In addition, because the initial straight walk corresponds to the unstable solution of the pitchfork bifurcation, it is unclear whether the robot will turn to the left or right, as shown in Fig.~\ref{curved}{\bf C} (the disturbance determines the turning direction and the robot continues walking straight unless disturbed).
Because the use of the optimal spring constant $\hat{k}_1$ does not necessarily guarantee the success of the approach to the target, we also used the supplementary turning controller in the leg yaw joints in the first module.

%この効果を明確にするために、$k_1=21\sim\hat{k}_1$に対して、補助制御なしの実験を行った。
%ただし、図と同じ条件で実験を行っている。
%補助制御を用いている場合は全て成功している。
%その一方で、用いない場合はターゲットに到達するものもあるが、予想通り曲線歩行に入るのに時間がかかったり、逆方向に旋回するなど、ターゲットに到達できていない。

\begin{figure}[t]
\begin{center}
	\includegraphics[width=88mm]{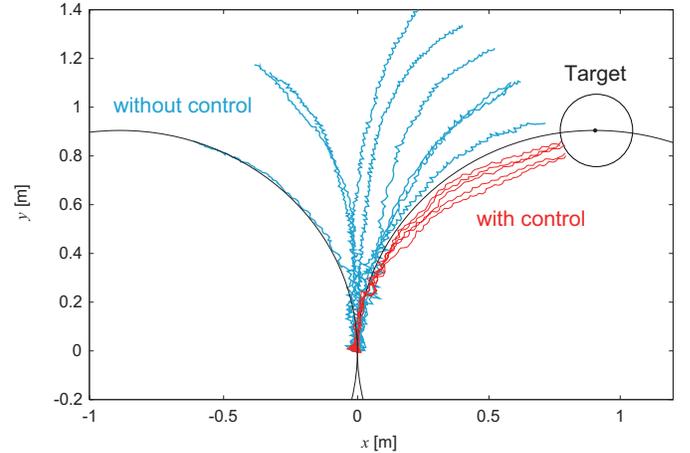}\vspace{0mm}

\caption{Comparison of the trajectory of the first module on the floor during the turning task with and without the supplementary turning controller. Nine trials without the supplementary controller and five trials with the controller (Fig.~\ref{turning}{\bf B}) are shown.
}
\label{without}
\end{center}
\end{figure}

To clarify the contribution of this supplementary controller, we also performed experiments without using the supplementary controller for $k_1=21$~Nmm/deg ($\sim\hat{k}_1$).
The experimental conditions were identical to those in Fig.~\ref{turning}{\bf B}.
Figure~\ref{without} compares the trajectory of the first module on the floor during the turning task with and without the supplementary controller.
In all the trials with the supplementary controller, the robot successfully approached the target.
In contrast, the robot without the supplementary controller failed in many trials, because the robot started walking in a straight line and took some time to converge to walking in a curved line, and it sometimes curved to the different direction from the target, as expected.

\subsection{Comparison with previous strategy}

%ピッチフォーク分岐を利用した最適な制御戦略の効果を調べるために、先行研究におけるホップ分岐を利用した実験を実施した。
%このとき、全てのヨー関節のばね定数は同じものを用い、次のばね定数を用いた(ホップ分岐の分岐点は次である)。
%ただし、ヨー関節のばね定数以外は図と同じ条件で実験を行っている。
%先行研究の結果と同様、分岐が生じる不安定化の近傍で最も旋回性能が良くなっているものの、今回のピッチフォークの方が剛性を適切に決めることで両方の評価指標においても優れた旋回性能を示した。

\begin{figure}[t]
\begin{center}
	\includegraphics[width=85mm]{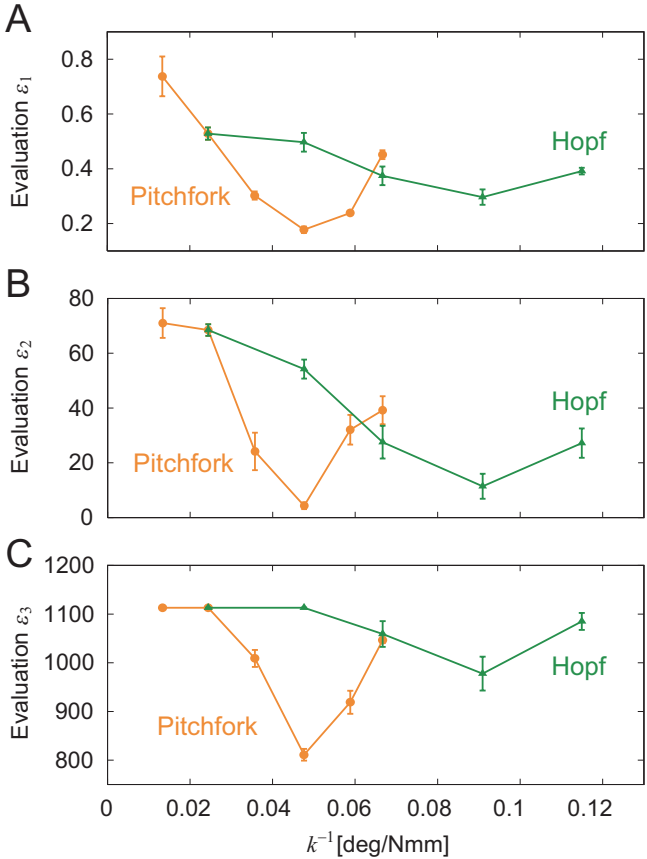}\vspace{0mm}

\caption{Comparison of evaluation criteria ({\bf A}) $\varepsilon_1$, ({\bf B}) $\varepsilon_2$, and ({\bf C}) $\varepsilon_3$ between the turning strategies based on pitchfork bifurcation and Hopf bifurcation. The data points and error bars correspond to the means and standard errors, respectively, of the results of five experiments.}
\label{comparison}
\end{center}
\end{figure}

To examine how the turning performance was improved by pitchfork bifurcation, we also performed experiments using the turning strategy based on Hopf bifurcation used in our previous work~\cite{bib_aoi2} and compared the performance.
For Hopf bifurcation, we used the same spring constant among the body-segment yaw joints and employed five spring constants ($k_i=8.7$, $11$, $15$, $21$, and $41$~Nmm/deg, $i=1,\dots,5$) to evaluate the turning performance for $k_i$, where the Hopf bifurcation point is about $k_i=18$~Nmm/deg ($1/\hat{k}_i=0.057$~deg/Nmm), as obtained in our previous work~\cite{bib_aoi2}.
The experimental conditions were identical to those in Fig.~\ref{turning}{\bf E} except for the spring constants of the body-segment yaw joints.
Figures~\ref{comparison}{\bf A}, {\bf B}, and {\bf C} compare the turning performance in terms of the criteria $\varepsilon_1$, $\varepsilon_2$, and $\varepsilon_3$, respectively, between the strategies based on pitchfork and Hopf bifurcations.
All criteria for Hopf bifurcation showed minimum values in the unstable region, as observed in our previous work~\cite{bib_aoi2}.
However, the minimum values of pitchfork bifurcation are lower than those of Hopf bifurcation for all criteria.
This means that the turning strategy based on pitchfork bifurcation created by tuning the body-axis flexibility is superior to that based on Hopf bifurcation, which was developed in our previous work~\cite{bib_aoi2}.

\section{Conclusion and discussion}

%この論文では、多脚ロボットの体軸柔軟性を変化させることで生じるピッチフォーク分岐により直線歩行が不安定化することを発見した。
%更に、この不安定化によって、体軸柔軟性に依存した曲率を持つ曲線歩行に遷移することを示した。
%このピッチフォーク分岐の特性を利用して、簡単な制御系を構築し、従来のホップ分岐よりもより機動的で効率の良い旋回を実現することができた。

In this study, we found that the straight walk of a many-legged robot with flexible body axis becomes unstable through pitchfork bifurcation when the body-axis flexibility is changed.
The straight walk transitioned into curved walk, whose curvature depended on the body-axis flexibility.
We developed a simple controller based on the pitchfork-bifurcation characteristics and demonstrated that the robot achieves high turning maneuver superior to the previous controller based on the Hopf bifurcation.

%進行方向の安定性は外乱からの抵抗や回復に関連し、機動性は意図的に進行方向を変える能力に関連する。
%これらの特性は大きく関わりを持っており、
%特に、不安定性は元の運動から逸脱させようとする駆動力を生み出すため機動性に貢献する。
%これらの特性は身体力学と環境との相互作用から決まるので、空気力学や流体力学による相互作用を介した移動ではこの関係が顕著に見られる。
%F16のような戦闘機は空力的に不安定になるように設計することで機動性を向上させており、不安定性を積極的に利用することは工学的にも有効な方法である。

Maneuverability is related to the ability to change movement direction.
When the movement direction is destabilized during locomotion, the instability provides driving forces to rapidly change the movement direction and thus enhances maneuverability.
Some military aircraft, such as the F-16, are designed to be aerodynamically unstable to enhance maneuverability~\cite{bib_avanzini1, bib_kwatny1}.
The use of dynamic instability is thus useful from an engineering viewpoint.

%不安定性を利用して移動運動の機動性を向上させる戦略は、生物的にも有効な手段である。
%その安定性は身体力学と環境との相互作用から決まるので、特に飛翔昆虫や水中動物など空気力学や流体力学による相互作用を介した移動では顕著となる。
%これは脚歩行においても顕著である。
%直立した脚を持つ四足動物のように質量中心が高いと、身体を左右に傾けて不安定性を引き起こして、旋回機動性を向上させる。
%しかしながら、横に張り出した脚を持つ爬虫類や節足動物のように質量中心が低いと、その歩行は平面的になる。
%そのため、身体を傾ける効果は低く、そのような旋回戦略はとれない。
%そのため、平面内での進行方向の安定性がより重要となる。
%実際、ゴキブリなどは、床反力の胴体に作用する場所を調整して平面内の直線歩行を不安定にし、機動性を上げると示唆されている。

The strategy of using movement direction instability to enhance maneuverability is also used by animals.
Because the instability is determined by the body dynamics during interaction with the environment, it is prominent in locomotion generated through aerodynamics and hydrodynamics, such as the locomotion of flying insects~\cite{bib_dickinson1, bib_parsons1, bib_taylor1} and sea animals~\cite{bib_fish1, bib_fish2, bib_webb1}.
It also appears in legged locomotion.
When the center of mass is high, as in mammals, whose legs are under the body, leaning the body to the left or right induces instability and helps turning~\cite{bib_courtine1, bib_schmitt2}.
However, when the center of mass is low, as in reptiles and arthropods, whose legs are away from the side of the body, locomotor behavior is almost two-dimensional because the center of mass moves in a horizontal plane.
Therefore, the effect of body leaning is small and thus such a turning strategy cannot be used, which implies that the stability of the walking direction in the horizontal plane becomes more crucial.
It has been suggested that cockroaches manipulate the position of the reaction forces from the floor entering the body to control the stability of a straight walk in a horizontal plane and that the straight walk instability helps their turning~\cite{bib_roctor1, bib_schmitt2}.

%これまでに、ヘビや魚型など推進に体軸の運動を利用する生物規範のロボットでは機動的な移動運動を実現している
%しかしながら、脚ロボットではまだそれほどの機動性は持つことは難しい。
%これは一つには、足の接地という地面との相互作用が断続的なものであるためである。
%この断続性は不整地踏破性などの利点ももたらすが、ロボットの転倒を容易に引き起こしてしまう。
%そのため、支持多角形やZMPに代表されるような力学的な指標を用いて、転倒回避のための制御開発に焦点が当てられており、機動性はよく調べられていなかった。
%それに対して、脚の数が増えると、転倒を回避できるようになる。
%その代わり、地面に拘束される脚の数が増え、機動性の障害となってしまう。
%更には、制御する自由度が増え、運動計画や制御が難しくなる。
%更に、一般的な多脚ロボットでは、脚だけでなく体節間もモータで制御するため、膨大な計算やエネルギーが必要になる。
%それに対して、本研究では、体節間が受動的なロボットを作成し、ロボットの力学に内在するピッチフォーク分岐に基づくシンプルな制御系を構築した。
%それは、体軸の動きを直接制御するのではなく、剛性を変えるだけなので、機動的と言うだけでなく、計算コストやエネルギーコストの削減にもつながった。
%ピッチフォーク分岐もホップ分岐も直線歩行の不安定性を引き起こすので、両方とも旋回機動性に貢献するが、ホップ分岐の蛇行への遷移に対して、ピッチフォーク分岐では曲線歩行に遷移するので、より旋回への貢献が大きい。
%更に、このような分岐が歩行力学系に導入できれば、蛇や魚など他のロボットの機動性の貢献にもつながる。

Various bioinspired robots that use their body axis for propulsion, such as snake robots~\cite{bib_astley1, bib_stamper1} and fish robots~\cite{bib_curet1, bib_maladen1, bib_sefati1}, have high maneuverability.
However, legged robots still have difficulty in achieving highly maneuverable locomotion.
This is partly because their interaction with the environment (i.e., foot contact with the ground) is intermittent due to the repetition of foot-contact and foot-off phases in leg movement.
Although this intermittency allows the traversal of diverse environments, it can make the robot lose balance.
Therefore, the control design of legged robots has focused on the avoidance of balance loss using dynamic criteria, such as a supporting polygon~\cite{bib_hirose1} and a zero-moment point~\cite{bib_vukobratovic1}, and maneuverability has not been well investigated.
Although increasing the number of legs prevents balance loss, it also increases the number of contact legs, which impedes maneuverability.
Moreover, the number of degrees of freedom to be controlled increases, making both motion planning and control difficult.
In addition, many-legged robots generally use actuators for controlling not only the leg joints but also body-segment joints~\cite{bib_takahashi1, bib_wei1}, which requires huge computational and energy costs.
In contrast, our robot has passive body-segment joints, which do not directly control the movement of the body axis, and instead determines the body-axis flexibility to induce a curved walk by the pitchfork bifurcation in the robot dynamics.
The generation of robot movements with dynamics rather than actuators is crucial for efficient locomotion~\cite{bib_collins1}, and our strategy greatly reduces the computational and energy costs.
Both the pitchfork and Hopf bifurcations induce the straight walk instability and thus contribute to the maneuverability.
However, because the pitchfork bifurcation causes a curved walk used for turning, unlike the Hopf bifurcation that causes body undulations, the pitchfork bifurcation makes a greater contribution.
Furthermore, when such bifurcation is introduced into the locomotion dynamics of other robots, such as snake and fish robots, through the mechanical and control design, it would contribute to improving their maneuverability.

%本研究では、一度の旋回のみに着目してピッチフォーク分岐の寄与を調べたが、連続した旋回などの複雑なシナリオにおいて、この制御系がどのように寄与するのかを明確にすることが重要である。
%ただし、連続した旋回では、旋回毎に、相対角度や距離などのターゲットの初期条件が異なる。
%ターゲットの初期条件に応じて旋回の曲率半径が変わるため、最適な旋回には体軸剛性も変える必要がある。
%これは現状の回転バネを用いたロボットでは不可能であり、体軸剛性を変化させる機構を導入するなど、より高度な制御が必要となる。
%これは今後の研究で詳細に明らかにしたい。
%四足や六足のロボットでは、CPGやセンサ生物規範型の制御を利用した、脚の運動調整を利用した旋回制御が提案されている。
%体軸の動特性に加えて、このような制御も統合することで、より高度な旋回が期待できる。

In the present study, we investigated the contribution of the pitchfork bifurcation to maneuverability in robot experiments where the robot performs turning only once on a flat floor.
It is important to clarify the contribution of the pitchfork bifurcation in more complex scenarios and environments in the future.
In particular, consecutive turnings are crucial for complex scenarios.
However, when we consider the experiments where the robot sequentially approaches multiple targets placed at different locations on the floor, that is, the robot performs multiple consecutive turnings, the initial conditions of the targets, such as the relative angle and distance, will differ in each turning.
Different initial conditions of the targets induce different radiuses of curvature for optimal turning, which require different body-axis flexibilities.
Because our robot uses torsional springs to determine the body-axis flexibility, it is impossible to change the flexibility during the experiment.
In future studies, we would like to improve our robot and controller, for example, by incorporating a variable flexibility mechanism in the body axis to conduct more complex tasks.
In addition, various turning strategies have been developed in quadruped and hexapod robots to modulate the leg movements for turning using bioinspired approaches based on central pattern generators and sensory systems~\cite{bib_arena1, bib_horvat1, bib_manoonpong1, bib_sprowitz1, bib_tsujita1, bib_zhu1} and optimization techniques~\cite{bib_degrave1, bib_roy1, bib_zhao1}.
We would like to improve our instability-based strategy in the body axis movement based on these strategies to enhance turning maneuverability of multi-legged robots in the future.

% or


\appendices
\section{Supplementary movies}

We recorded two supplementary movies to show the pitchfork bifurcation of a straight walk and turning performance in the robot experiments:

\begin{enumerate}[\textrm{Movie~}1:]
\item Straight walk, curved walk with a small curvature, and curved walk with a large curvature using a large spring constant, small spring constant, and very small spring constant, respectively, for the torsional spring in yaw joint 1.
\item Unsuccessful approaches using a spring constant larger and smaller than the optimal value for yaw joint 1, and successful approach using a spring constant close to the optimal value.
\end{enumerate}

\section{Supplementary turning control by leg joints} \label{Append_controller}

%最適な旋回戦略は、初期の位置関係に応じたフィードフォワード制御である。
%更に、ピッチフォーク分岐では左右どちらに旋回するかは初期のロボットの状態に依存する。
%ターゲット追従を保証するために、先行研究で構成したフィードバックに基づく旋回制御系も併用する。
%詳細には、測域センサで計測した相対角と先頭の脚ヨー関節を用いる。
%1歩ごとに($0\le t<T$, $T$：歩行周期)，先頭モジュールの脚のヨージョイント角の目標値を次のように決定した。
%ここで、$t_{start}$は遊脚時間の40\%，$t_{end}$遊脚時間の80\%としている。
%これは，脚ヨージョイント毎に，一歩の間に、最大旋回角を$5^\circ$として、遊脚相の間にターゲットの方に進行方向を変えるものである．
%更に、旋回タスクにおいて、脚よ−ジョイントの最大角を5度と制限している。
%このフィードバック制御は、ピッチフォーク分岐で構成される最適な円軌道に追従させるわけではない。
%そうではなく、ターゲットの方向に応じて先頭が進行方向を決め、モジュール間の受動的な結合を通して先頭モジュールに追従することで、旋回タスクを実現させようというものである。
%そのため、$k_1\ne\hat{k}_1$でもターゲットに近づけることができる。

The optimal turning strategy is feedforward, depending only on the initial relative position between the robot and target.
In addition, the direction in which the robot turns (left or right) depends on the initial robot conditions because of the property of pitchfork bifurcation.
To guarantee a successful approach to the target, we used a supplementary feedback-based turning controller, which was developed in our previous work~\cite{bib_aoi2}.
Specifically, we used the relative target angle $\psi$ of the first module measured by the laser range scanner and the leg yaw joints $\psi_1$ and $\psi_2$ of the first module.
We determined the desired angles $\hat{\psi}_1$ and $\hat{\psi}_2$ of $\psi_1$ and $\psi_2$ for each gait cycle [$t_i^n\le t<t_i^n+T$, where $t=t_i^n$ is the time when the desired leg tip is at the PEP for the $n$th gait cycle, and $T$ is the gait cycle duration ($=0.6$~s)] using
\begin{eqnarray}
\hat{\psi}_i(t)&\hspace{-3mm}=&\hspace{-3mm}\left\{\hspace{-1mm}\begin{array}{ll}
\hat{\psi}_i(t_i^n) & \hspace{-1mm}t_i^n\le t<t_i^n+t_{\scriptsize{\mbox{start}}} \\
\hat{\psi}_i(t_i^n)+\Delta_i\displaystyle\frac{t-t_i^n-t_{\scriptsize{\mbox{start}}}}{t_{\scriptsize{\mbox{end}}}-t_{\scriptsize{\mbox{start}}}} & \hspace{-1mm}t_i^n+t_{\scriptsize{\mbox{start}}}\le t\le t_i^n+t_{\scriptsize{\mbox{end}}} \\
\hat{\psi}_i(t_i^n+t_{\scriptsize{\mbox{end}}}) & \hspace{-1mm}t_i^n+t_{\scriptsize{\mbox{end}}}<t<t_i^n+T
\end{array}\right. \nonumber \\
\Delta_i&\hspace{-3mm}=&\hspace{-3mm}\left\{\hspace{-1mm}\begin{array}{l}
\psi(t_i^n+t_{\scriptsize{\mbox{start}}})-\hat{\psi}_i(t_i^n+t_{\scriptsize{\mbox{start}}}) \\
\hspace{20mm} |\psi(t_i^n+t_{\scriptsize{\mbox{start}}})-\hat{\psi}_i(t_i^n+t_{\scriptsize{\mbox{start}}})|<5^\circ \\
5^\circ \hspace{17mm} \mbox{otherwise}
\end{array}\right. \nonumber
\end{eqnarray}
where $t_{\scriptsize{\mbox{start}}}$ and $t_{\scriptsize{\mbox{end}}}$ were set to $40$\% and $80$\%, respectively, of the duration of the half elliptical curve of the leg tip trajectory ($=0.12$ and $0.23$~s) determined experimentally.
This controller means that each leg changed its yaw direction toward the target only during the swing phase with $5^\circ$ of the maximum turning angle for one gait cycle.
We also limited the maximum angle of the leg yaw joint to $5^\circ$ during the turning task.
This supplementary control did not aim to make the robot follow the optimal curved path generated by the turning strategy using pitchfork bifurcation.
Instead, it was designed so that the first module modulated the walking direction based on the target direction, which solves the problems related to the feedforward property of the optimal turning strategies and the turning direction due to initial robot conditions, and furthermore allows the robot to approach the target even when $k_1\ne\hat{k}_1$.

\section*{Acknowledgment}

This study was supported in part by JSPS KAKENHI Grant Numbers JP17H04914, JP19KK0377, and JP20H00229; JST FOREST Program Grant Number JPMJFR2021; and the Inamori Foundation.

\ifCLASSOPTIONcaptionsoff
  \newpage
\fi

\end{document}